\documentclass[journal,dvipsnames]{IEEEtran}

\usepackage[T1]{fontenc}
\usepackage{amsfonts}
\usepackage{amsmath}
\usepackage{amssymb}
\usepackage{calc}
\usepackage{etoolbox}
\usepackage{graphicx}
\usepackage{multirow}
\usepackage{pgfplots}
\usepackage{tikz}
\usepackage{xcolor}
\usepackage{xparse}

\DeclareMathOperator*{\nll}{NLL}

\DeclareMathOperator{\softmax}{softmax}
\DeclareMathOperator{\softplus}{softplus}

\def\MyPgfPlotsCsvPath{./csv}
\def\MyPgfPlotsCsvSep{comma}

\usetikzlibrary{arrows.meta}
\usetikzlibrary{matrix}
\usetikzlibrary{positioning}

\tikzstyle{signal}=[-{latex},gray, text=black,line width=0.8pt,shorten >=1pt]
\tikzstyle{inv signal}=[{latex}-,gray, text=black,line width=0.8pt,shorten <=1pt]
\tikzstyle{net node}=[circle, minimum width=2em, align=center, line width=0.8pt]
\tikzstyle{input node}=[net node, fill=blue!20, draw=black]
\tikzstyle{hidden node}=[net node, draw=black]
\tikzstyle{output node}=[net node, fill=orange!20,draw=black]

\tikzstyle{annot} = [align=center, font=\small]

\def\hideNode{hidden}
\def\noName{no_name}
\newif\iftikzAnnAnnotate
\pgfkeys{%
  tikzANN/.is family,
  tikzANN/.cd,
  node dist/.store in = \tikzAnnNodeDist,
  layer dist/.store in = \tikzAnnLayerDist,
  pin dist/.store in = \tiksAnnPinDist,
  annotate/.is if = tikzAnnAnnotate
}
\newcommand{\tikzANN}[4][]{%
  \pgfkeys{%
    tikzANN/.cd,
    node dist = 1cm,
    layer dist = 1cm,
    pin dist = 10pt,
    annotate=true,
    #1
  }
  \ifnum0<0#2\relax
    \def\inputneurons{1,...,#2}
  \else
    \def\inputneurons{#2}
  \fi
  \def\hidenneurons{#3}
  \def\outputneurons{#4}
  \begin{tikzpicture}[%
      remember picture,
      every pin edge/.style={signal}
    ]
    \matrix [%
      matrix of nodes,
      column sep=\tikzAnnLayerDist,
      align=center,
      nodes={anchor=center},
      ampersand replacement=\&
    ] {
      \begin{tikzpicture}[remember picture]
        \foreach [count=\c] \y / \name in \inputneurons{
          \ifx\name\hideNode
            \node at (0,-\y*\tikzAnnNodeDist) {\raisebox{1.2ex}{$\vdots$}};
          \else
            \node[input node]
            (I\c) at (0,-\y*\tikzAnnNodeDist) {%
              \ifx\name\y%
                $x_\c$%
              \else%
                \name%
              \fi%
            };
          \fi
        }
      \end{tikzpicture}
      \&
      \begin{tikzpicture}[remember picture]
        \foreach [count=\c]  \y in {1,...,\hidenneurons}
        \node[hidden node]
        (H\c) at (0,-\y*\tikzAnnNodeDist) {};
      \end{tikzpicture}
      \&
      \begin{tikzpicture}[remember picture]
        \foreach [count=\c] \y / \name / \pin in \outputneurons{
          \ifx\name\hideNode
            \node (O\c) at (0,-\y*\tikzAnnNodeDist) {\raisebox{1.2ex}{$\vdots$}};
          \else
            \ifx\name\y%
              \ifx\pin\name
                \def\name{$y_\c$}%
              \else
                \tikzset{
                  output node/.append style={%
                      pin={[pin distance=\tiksAnnPinDist]right:\pin}
                    }
                }
                \def\name{$y_\c$}%
              \fi%
            \else\ifx\name\noName%
                \def\name{}%
              \else
                \ifx\pin\name
                  \tikzset{
                    output node/.append style={%
                        pin={[pin distance=\tiksAnnPinDist]right:\pin}
                      }
                  }
                  \def\name{$y_\c$}%
                \else
                  \tikzset{
                    output node/.append style={%
                        pin={[pin distance=\tiksAnnPinDist]right:\pin}
                      }
                  }
                \fi%
              \fi
            \fi
            \node[output node]
            (O\c) at (0,-\y*\tikzAnnNodeDist) {\name};
          \fi
        }
      \end{tikzpicture}
      \\
    };

    \foreach  [count=\cs] \source / \name in \inputneurons{
      \ifx\name\hideNode\else
        \foreach  [count=\cd] \dest in {1,...,\hidenneurons}
        \draw[signal] (I\cs) -- (H\cd);
      \fi
    }
    \foreach  [count=\cd] \dest / \n / \hidden in \outputneurons{
      \ifx\hidden\hideNode\else
        \foreach  [count=\cs] \source in {1,...,\hidenneurons}
        \draw[signal] (H\cs) edge (O\cd);
      \fi
    }

    \iftikzAnnAnnotate
      \node[annot] at ($(current bounding box.north -| H1) + (0, 0.5)$) (hl) {Hidden\\layer};
      \node[annot] at (I1 |- hl) {Input\\layer};
      \node[annot] at (O1 |- hl) {Output\\layer};
    \fi
  \end{tikzpicture}
}

\pgfplotsset{compat=newest}
\usepgfplotslibrary{dateplot}
\usepgfplotslibrary{fillbetween}
\usepgfplotslibrary{groupplots}

\pgfplotsset{
    plot style/.style={%
        legend pos=outer north east,
        legend style={%
            draw=none,
        },
        legend cell align=left,
        axis x line=bottom,
        axis y line=left,
        enlarge y limits=true,
        enlarge x limits=.01,
    },
    forecast plot style/.style={%
        plot style,
        date coordinates in=x,
        xtick distance = 0.25,
        xticklabel={\hour:\minute}
    },
    label style={font=\small},
}

\NewDocumentCommand{\dateTimeLoadForecastConfidenceAddPlot}{O{99,60} O{q500/Median/RoyalBlue,y/Observation/Orange} m}{%
  \foreach \ci in {#1}{
      \pgfmathtruncatemacro{\qLow}{(100 - \ci) * 10 / 2}
      \pgfmathtruncatemacro{\qHigh}{\ci * 10 + \qLow}
      \pgfmathsetmacro{\opacity}{(100-\ci)/ 100 + 0.2}
      \addplot [
        draw=none,
        forget plot,
        name path=qLow
      ] table [
          x=date_time,
          y=q\qLow,
          col sep=\MyPgfPlotsCsvSep,
          search path=\MyPgfPlotsCsvPath
        ] {#3.csv};
      \addplot [
        draw=none,
        forget plot,
        name path=qHigh
      ] table [
          x=date_time,
          y=q\qHigh,
          col sep=\MyPgfPlotsCsvSep,
          search path=\MyPgfPlotsCsvPath
        ] {#3.csv};

      \edef\tmp{
        \noexpand\addplot[
          RoyalBlue,
          opacity=\opacity
        ] fill between[of=qLow and qHigh];
        \noexpand\label{pgfplots:#3_\ci}
        \noexpand\addlegendentry{$CI_{\ci\,\%}$}
      }
      \tmp
  }

  \foreach \y/\n/\c in {#2}{
    \edef\tmp{
      \noexpand\addplot [
        \c
      ] table [
          x=date_time,
          y=\y,
          col sep=\MyPgfPlotsCsvSep,
          search path=\MyPgfPlotsCsvPath
        ] {#3.csv};
        \noexpand\label{pgfplots:#3_\n}
        \noexpand\addlegendentry{\n}
    }
    \tmp
  }
}
\NewDocumentCommand{\samplePlot}{o O{99} m m}{%
  \begin{tikzpicture}
    \begin{axis}[%
        forecast plot style,
        ylabel style={align=center},
        ymin=0,%
        #1
      ]
      \foreach \n in {0,...,#4} {%
          \addplot [gray, opacity=.75] table [%
              x=date_time,%
              y=s\n,%
              col sep=\MyPgfPlotsCsvSep,
              search path=\MyPgfPlotsCsvPath,
              \ifnum\n>0%
                forget plot%
              \fi
            ] {#3.csv};
        }
      \label{pgfplots:#3_Samples}
      \addlegendentry{Samples}
      \dateTimeLoadForecastConfidenceAddPlot[#2][{q500/Median/RoyalBlue,y/Observation/{Orange,thick}}]{#3}
    \end{axis}
  \end{tikzpicture}
}

\usepackage[acronym]{glossaries-extra}
\setabbreviationstyle[acronym]{nolong-short}

\newacronym{1DCNN}{1DCNN}{dilated 1D-Convolution neural network}
\newacronym{BNF}{BNF}{Bernstein-Polynomial Normalizing Flow}
\newacronym{CNN}{CNN}{Convolutional Neural Networks}
\newacronym{CPD}{CPD}{Conditional Probability Distribution}
\newacronym{CRPS}{CRPS}{Continuous Ranked Probability Score}
\newacronym{NCRPS}{NCRPS}{normalized~\gls{CRPS}}
\newacronym{ELU}{ELU}{Exponential Linear Unit}
\newacronym{FC}{FC}{fully connected neural network}
\newacronym{GMM}{GMM}{Gaussian Mixture Models}
\newacronym{GM}{GM}{Gaussian Model}
\newacronym{LSTM}{LSTM}{Long Short-term Memory}
\newacronym{LV}{LV}{low-voltage}
\newacronym{MAE}{MAE}{Mean Absolute Error}
\newacronym{MQS}{MQS}{Mean Quantile Score}
\newacronym{NMQS}{NMQS}{normalized~\gls{MQS}}
\newacronym{MSE}{MSE}{Mean Squared Error}
\newacronym{NF}{NF}{Normalizing Flow}
\newacronym{NLL}{NLL}{Negative Logarithmic Likelihood}
\newacronym{NN}{NN}{Neural Networks}
\newacronym{QR}{QR}{Quantile Regression}
\newacronym{ReLU}{ReLU}{Rectified Linear Unit}
\newacronym{ECDF}{ECDF}{Empirical Cumulative Density Function}

\usepackage{multirow}
\usepackage{booktabs} 
\usepackage[
  style=ieee,
  dashed=false,
  hyperref=true,
  url=true,
  backend=biber,
  natbib=true
] {biblatex}
\graphicspath{{./gfx/}}

\usepackage{hyperref}
\usepackage[nameinlink,capitalize]{cleveref}

\def\ieeedoi{Digital Object Identifier: \href{doi.org/10.1109/TSG.2023.3254890}{10.1109/TSG.2023.3254890}}

\title{Short-Term Density Forecasting of Low-Voltage Load using Bernstein-Polynomial Normalizing~Flows}

\author{Marcel Arpogaus,
  Marcus Voss,
  Beate Sick,
  Mark Nigge-Uricher
  and Oliver Dürr
  \thanks{M. Arpogaus and O. Dürr are with HTWG Konstanz - University of Applied Sciences,
    e-mail: marcel.arpogaus@htwg-konstanz.de}
  \thanks{M. Voss is with TU Berlin and Birds on Mars GmbH}
  \thanks{B. Sick is with EBPI, University of Zurich \& IDP, Zurich University of Applied Sciences}
  \thanks{M. Nigge-Uricher is with Bosch.IO GmbH}
  \thanks{Manuscript received April 29, 2022; revised Jan 23, 2023; accepted Mar 3, 2023; Date of Publication Mar 11, 2023; date of current version Mar 7. 2023. For information on obtaining reprints of this article, please send  e-mail to: reprints@ieee.org. \ieeedoi}}

\begin{document}

\markboth{\textcopyright 2023 IEEE. \ieeedoi}
{Arpogaus \MakeLowercase{\textit{et al.}}: Short-Term Density Forecasting of Low-Voltage Load using Bernstein-Polynomial Normalizing Flows}

\maketitle

{\bfseries\textcopyright{} 2023 IEEE. Personal use of this material is permitted. Permission from IEEE must be obtained for all other uses, in any current or future media, including reprinting/republishing this material for advertising or promotional purposes, creating new collective works, for resale or redistribution to servers or lists, or reuse of any copyrighted component of this work in other works.}

\medskip

\begin{abstract}
  The transition to a fully renewable energy grid requires better forecasting of demand at the low-voltage level to increase efficiency and ensure reliable control.
  However, high fluctuations and increasing electrification cause huge forecast variability, not reflected in traditional point estimates.
  Probabilistic load forecasts take uncertainties into account and thus allow more informed decision-making for the planning and operation of low-carbon energy systems.
  We propose an approach for flexible conditional density forecasting of short-term load based on Bernstein polynomial normalizing flows, where a neural network controls the parameters of the flow. In an empirical study with 3639 smart meter customers, our density predictions for 24h-ahead load forecasting compare favorably against Gaussian and Gaussian mixture densities.
  Furthermore, they outperform a non-parametric approach based on the pinball loss, especially in low-data scenarios.
\end{abstract}


\IEEEpeerreviewmaketitle

\section{Introduction} \label{sec:introduction}

\subsection{Motivation}
On the path to a sustainable energy supply, the take-up of renewable and distributed energy resources transforms the electric energy system into a more decentralized system. This increases the role of~\glsxtrfull{LV} grids that typically make up the largest part of distribution systems but are still the least monitored and controlled. Regulations such as the German ``Klimaschutzgesetz'' or ``UK Green Industrial Revolution: 10 Point Plan'' and their consequent funding programs promote the electrification of mobility, heating, and industrial process, leading to rapid changes in electricity generation and consumption -- to the extent that they already cause significant changes in load curves and power flows in the distribution network~\cite{Heymann2019}.

Hence, accurate local short-term load and generation forecasts at the~\gls{LV} level ranging from minutes to days ahead are becoming essential for grid operators, utilities, building- and district operators, and the customers themselves to make informed decisions within many applications. Such applications include, for instance, peak load reduction~\cite{Rowe2014} and voltage control~\cite{Zufferey2020}. Accurate load forecasts can also be used for grid state estimation~\cite{Hermanns2020}. 
They can also be used for anomaly detection to increase resilience for the grid~\cite{Fadlullah2011} or to detect energy theft~\cite{Fenza2019}. Short-term forecasts can further inform participants of local and peer-to-peer energy markets~\cite{Morstyn2018}, real-time pricing schemes~\cite{He2019a} or flexibility applications~\cite{Ponocko2018} that are emerging with the energy transition. See~\cite{Haben2021} and \cite{Wang2019d} for reviews on more applications.

There are a plethora of load forecasting approaches that have been proposed on the aggregated and system-level. While most approaches are focused on point forecasts, a trend towards probabilistic approaches can be seen (see, for instance, as reviewed in~\cite{VanderMeer2018, Hong2020, Hong2016a}).
While usual point forecasts model the expected value, probabilistic forecasts model the distribution, enabling more informed decision-making by considering uncertainties. This makes probabilistic forecasts especially relevant for the~\gls{LV} level, as they can better handle the higher volatility present at this level when compared to the system-level~\cite{Hong2020, Haben2021}.

\subsection{Related Work}

Probabilistic forecasting at the~\gls{LV} level is challenging, as the time series are volatile, multivariate, and the marginal distributions are typically skewed and multi-modal~\cite{Anvari2020}.
Generally, uncertainty in load forecasts can be modelled as either simple prediction intervals~\cite{Kodaira2020}, quantile estimates~\cite{Wang2019b,Gerossier2018, Elvers2019, Chen2019a} (providing estimates for a fixed set of quantiles), full continuous distributions~\cite{Pinto2017, Arora2016} or scenarios~\cite{Khoshrou2019}, see~\cite{Haben2021} for a review.

Further, the methods applied can be distinguished between statistical and time series models that often rely on simple parametric assumptions and
models from the machine learning or deep learning domain, which can model complex dependencies at the cost of interpretability.
Typical statistical models, capable of forecasting intervals and quantiles, are, for example, kernel regression~\cite{Chaouch2015}, Gaussian modelling~\cite{Kodaira2020} and additive quantile regression~\cite{BenTaieb2016}. Statistical models for the continuous distribution are the ARMA-GARCH~\cite{Bikcora2018} models.
In deep learning, it is straightforward to model the full density distribution, with complex dependencies on explanatory variables in the input data. In this case, the distribution parameters are modelled by the last layer of a neural network. The complete network can then simply be trained with state of the art stochastic gradient descent and the standard loss function defined by the maximum likelihood principle~\cite{Durr2020}. So far, only rather simple parametric DL approaches have been proposed in the LV context, which make strong assumptions about the underlying distribution, such as
a single Gaussian distribution~\cite{Haben2019a, Salinas2020} or~\glsxtrfull{GMM}~\cite{Vossen2018}.
The modelling for more complex distributions in the context of~\gls{LV} level load forecasts has been done so far mostly indirectly via quantile regression for different~\gls{NN} architectures: \glsxtrfull{NN}~\cite{Wang2019c,Zufferey2020,Zhang2020a}, Residual Networks~\cite{Chen2019a}, \glsxtrfull{LSTM}~\cite{Yang2020, Wang2019b} and~\glsxtrfull{CNN}~\cite{Elvers2019}. These methods cannot be minimized using the standard maximum likelihood principle, and instead rely on an ad-hoc loss like the pinball loss. Further, care has to be taken to avoid quantile crossing.

\subsection{Contribution of this Work}

Here, we propose a flexible method which fits directly into the deep learning framework. It is based on
\emph{\glsxtrfullpl{NF}} flexible parameterized transformations from a simple distribution (e.g., Gaussian) to an arbitrary complex one (see~\cite{Papamakarios2021} for an overview).
Up to now, \gls{NF} models have gained the most attention in applications where complex high-dimensional unconditioned distributions $p_y(\mathbf{y})$ are modelled, i.\,e. image generation~\cite{Kingma2018} or speech synthesis~\cite{Oord2018}. Probabilistic regression models based on~\gls{NF}, modelling the~\gls{CPD} $p_y(y|\mathbf{x})$ have gained little attention.
However, recent research extended~\gls{NF} to conditional density estimation with very promising results~\cite{Rothfuss2020, Trippe2018, Sick2021} and applied it to time-series forecasting~\cite{Rasul2020} and scenario generation of
residential loads~\cite{Zhang2020, Ge2020}.

\emph{We propose an approach based on~\glsxtrfullpl{BNF} for short-term density forecasting of~\gls{LV} loads}, which have been used in the statistics community for a while~\cite{Hothorn2018} and were brought to deep learning by~\cite{Sick2021}.
Compared to existing methods, \gls{BNF} allow full density forecasts without strong parametric assumptions on the distribution and flexible modelling of explanatory variables. This makes the approach especially suitable for the challenging~\gls{LV} level.
Our empirical results demonstrate that the proposed method is capable of fitting a single model for multiple customers, rather than requiring a separate model for each individual customer.
Moreover, our results indicate that models using deep artificial neural networks to estimate density parameters can generalize well to previously unseen but similar situated households.

This work extends an earlier preliminary version~\cite{Arpogaus2021}, with an in-depth discussion of the developed method and a more rigorous comparison with the other methods.
The robustness has been confirmed by the stable test performances of 10 fits of the model, each trained on the same train data after random initialization.

In~\cref{sec:method} we give a brief introduction to~\glspl{NF} and describe our~\gls{BNF} approach for short-term density forecasting of~\gls{LV} loads and highlight the differences to previous~\gls{BNF} implementations.
In~\cref{sec:study} we use public data from $3,639$ smart meter customers of the CER dataset~\cite{CER2012} to compare two different~\gls{NN} architectures combined with four parametric and non-parametric density estimation methods in 24h-ahead forecasting.
Finally, we conclude this study in~\cref{sec:conclusion}.

\section{Normalizing Flows using Bernstein-Polynomials} \label{sec:method}

We tackle the load forecasting problem in the framework of deep probabilistic regression. For the covariates $\mathbf{x}$ such as lagged power consumption at earlier time steps, holiday indicator, or calendar variables, the~\glsxtrfullpl{CPD} $p_y(y_t|\mathbf{x})$, of the electric load at time-step $t$ are predicted (\cref{fig:model-scheme}).
For notational convenience, we drop the subscript $y$ and $t$ when there is no ambiguity.

\subsection{Background on Normalizing Flows (NF)}\label{sec:nf}
The main idea of~\gls{NF} is to fit a parametric bijective conditional function $z=f(y|\theta(\mathbf{x}))$ that transforms between a possibly complex conditional target distribution $p_y(y|\mathbf{x})$ and a simple distribution $p_z(z)$, often $p_z(z)=N(0,1)$. The change of variable formula allows us to calculate the probability $p_y(y|\mathbf{x})$ from the simple probability $p_z(z)$ as follows:
\begin{equation}
  p_y(y|\mathbf{x}) = p_z\left(f(y)|\theta(\mathbf{x})\right) \left|\det\nabla{f}(y|\theta(\mathbf{x}))\right|
  \label{eq:prob}
\end{equation}
With the Jacobian determinant $\det\nabla{f(y|\theta(\mathbf{x}))}$ ensuring that the probability integrates to one after the transformation (hence the name normalizing flow).
Sampling from the learned \emph{data distribution} $p_y$ is then achieved by first sampling $z$ from the simple \emph{base distribution} $p_z$ and passing it through the \emph{inverse transformation} $f^{-1}(z|\theta(\mathbf{x}))$ to obtain $y$.
This leads to the two main properties the transformation functions must satisfy: i) $f$ must be \emph{invertible}, usually guaranteed by restricting the transformation function to be strictly monotone, and ii) both $f$ and $f^{-1}$ must be \emph{differentiable}~\cite{Papamakarios2021}.

The parameters $\theta$ of the bijective transformation functions $f$ can then be tuned by maximizing the likelihood $\mathcal{L}$ of the samples $y$, observed under condition $\mathbf{x}$ via~\cref{eq:prob}.
For numerical reasons, it is common practice to minimize the negative~\glsxtrfullpl{NLL} function instead
\begin{equation}\label{eq:nll}
  \nll = - \sum_{y|\mathbf{x} \in \mathcal{D}}\log \left(p_y(y|\mathbf{x},\theta)\right)
\end{equation}
\subsection{Probabilistic regression with Bernstein Flows}

Most of the recent NF methods in the Machine Learning literature construct the flow from a combination of $K$ simple transformation functions $f_i$, to compose more expressive transformations $f(z)= f_K \circ f_{K-1} \circ \dots \circ f_1(y)$~\cite{Papamakarios2021}.
Chaining many of those simple transformations can result in a flexible bijective transformation.
Alternatively, there are approaches that use a single flexible transformation, such as sum-of-squares polynomials~\cite{Jaini2019}, or splines~\cite{Durkan2019a, Durkan2019b}.
Our approach benefits from Bernstein polynomials introduced recently in the statistics community~\cite{Hothorn2018} and combined with~\gls{NN} in~\cite{Sick2021, Ramasinghe2021}.
Compared to other methods~\glsxtrfull{BNF} have several advantages, like
1) robustness against initial and round-off errors,
2) higher interpretability,
3) a theoretical upper bound for the approximation error,
4) ability to increase the flexibility at no cost to the training stability.
See~\cite{Ramasinghe2021} for an in-depth theoretical analysis of the framework.

Our implementation extends the work of~\citet{Sick2021} but has some significant improvements to enhance convergence and allow a stable inversion of the flow from the latent variable $z$ to the observed $y$, which is needed, i.e., for sampling new load profiles from the learned distributions (\cref{fig:sampling}).

\begin{figure}[ht!]
  \centering
  \includegraphics[
    keepaspectratio,
    width=0.9\columnwidth,
  ]{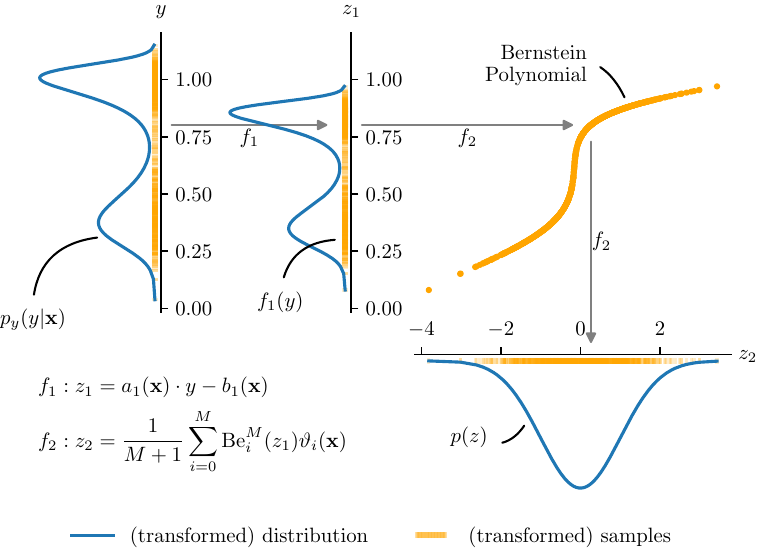}
  \caption{The~\gls{NF} $f = f_2 \circ f_1$ transforms the bimodal distribution $p(y|\mathbf{x})$ in the upper-left side to the standard Gaussian $p(z)$ shown in the lower-right side. The dependence on the covariates $\mathbf{x}$ stems from the x-dependence of the~\gls{NN} controlled parameters $a_1, b_1$ and $\vartheta_0,\ldots,\vartheta_M$ (\cref{fig:NN-para}). $f_2 $ uses Bernstein polynomials for maximal flexibility.}%
  \label{fig:method}
\end{figure}

The complete~\gls{NF} $f = f_2 \circ f_1$ consists of two functions, $f_1$ and $f_2$, as shown in~\cref{fig:method}.
The first transformation $f_1: z_1=a_1(\mathbf{x})\cdot{y}-b_1(\mathbf{x})$, allows transforming $y$ into the domain $[0,1]$ of the Bernstein polynomial $f_2$ of order $M$ defined as:
\begin{equation}
  f_2(z_1) = \frac{1}{M+1}\sum_{i=0}^M \operatorname{Be}_i(z_1) \vartheta_i
\end{equation}
Generated by the $M+1$ densities of the Beta distribution $\operatorname{Be}_i(z) = f_{i+1, M-i+1}(z)$, multiplied with the corresponding Bernstein coefficients $\vartheta_0,\ldots,\vartheta_M$.
Bernstein polynomials were first introduced by Bernstein~\cite{Bernstein1912} to prove the Weierstrass approximation theorem. This theorem states that every continuous function on any fixed interval can be approximated arbitrarily well by a polynomial with sufficient order. Bernstein introduced the Bernstein polynomials for a constructive proof by showing that they can approximate every function in $f: z \in [0, 1] \rightarrow \mathbb{R}$ for $M \rightarrow \infty$ (see~\cite{Farouki2012} for details of the proof and further properties of the Bernstein polynomials).
Higher degree Bernstein polynomials increase the expressiveness with no cost to the training stability~\cite{Hothorn2018, Ramasinghe2021}.
Empirically, $M \gtrsim 10$ polynomials are often sufficient in typical regression settings~\cite{Hothorn2018}.

The Bernstein polynomials are bounded between $z \in [0,1]$.
In~\cite{Sick2021} this has been ensured by a sigmoid function, but when evaluating the inverse flow, we found some serious drawbacks in numerical stability.
Hence, we decided to remove the sigmoid function and instead do a linear extrapolation of $f_2$ for $z \notin [0,1]$.
Moreover, \cite{Sick2021} used a second affine transformation $f_3$.
Since we could not find a beneficial effect (w.r.t. NLL) of $f_3$, we removed it completely.

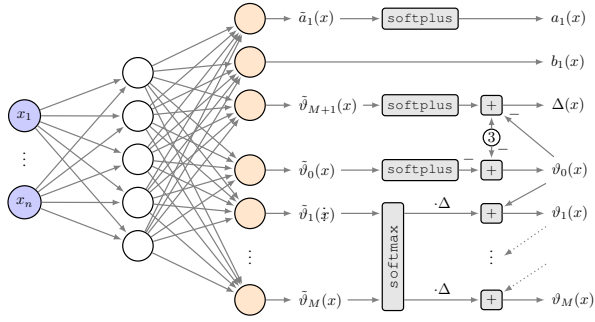
\begin{figure}[h]
  \resizebox{\columnwidth}{!}{
\pgfkeysifassignable{/tikz/external/export next}{%
  \tikzset{external/export next=false}
}{}
\usetikzlibrary{fit}
\begin{tikzpicture}[
    node distance=1pt,
    remember picture,
    activation/.style={
        rectangle,
        rounded corners = 2pt,
        font=\small,
        fill=black!10,
        draw=black,
        line width=0.8pt
      },
    constant/.style={
        circle,
        draw=black,
        align=center,
        line width=0.8pt,
        inner sep=0.1em
      }
  ]
  \node {%
    \tikzANN[layer dist = 1.5cm, annotate = false]
    {{1},{2}/hidden,3/$x_n$}
    {5}
    {
      1/no_name,2/no_name,%
      3/no_name,4.5/no_name,5.5/no_name,6.5/hidden,7.5/no_name%
    }
  };
  \draw[signal] (O1)       -> +(1,0)         node [right] (ta1) {$\tilde{a}_1(x)$};
  \draw[signal] (ta1)      -> +(1.5,0)       node [right, activation] (ac1) {\texttt{softplus}};
  \draw[signal] (ac1.east) -> +(2,0)         node [right] (a1) {$a_1(x)$};

  \draw[signal] (O2)       -> (O2-|a1.west)  node [right] (b1)  {$b_1(x)$};

  \draw[signal] (O3)       -> +(1,0)         node [right] (gM)  {$\tilde\vartheta_{M+1}(x)$};
  \draw[signal] (gM)       -> (gM-|ac1.west) node [right, activation] (spM) {\texttt{softplus}};
  \draw[signal] (spM.east) -> +(0.5,0)       node [right, activation] (pM) {$+$};
  \draw[signal] (pM.east)  -> +(1,0)         node [right] (delta)  {$\Delta(x)$};

  \draw[signal] (O4)       -> +(1,0)         node [right] (g0)  {$\tilde\vartheta_{0}(x)$};
  \draw[signal] (g0)       -> (g0-|ac1.west) node [right, activation] (sp0) {\texttt{softplus}};
  \draw[signal] (sp0.east) ->                node[pos=1,above left] {$-$} +(0.5,0)         node [right, activation] (p0) {$+$};
  \draw[signal] (p0.east)  -> +(1,0)         node [right] (t0)  {$\vartheta_{0}(x)$};

  \node[constant, below=0.3 of pM] (std) {$3$};
  \draw[signal] (std)      -> (pM);
  \draw[signal] (std)      ->                node[pos=1,above right] {$-$} (p0);

  \draw[signal] (t0)       ->                node[pos=1, right] {$-$} (pM);

  \draw[signal] (O5)       -> +(1,0)         node [right] (g1)  {$\tilde\vartheta_{1}(x)$};
  \draw[signal] (O7)       -> +(1,0)         node [right] (gM-1)  {$\tilde\vartheta_{M}(x)$};

  \node[
    activation,
    fit={(g1-|ac1.west) (gM-1-|ac1.west)},
    anchor=west,
    inner sep=0.7em,
    label={[rotate=90]center:\texttt{softmax}}
  ] (sm) {};
  \draw[signal] (g1)            -> (g1-|sm.west);
  \draw[signal] (gM-1)          -> (gM-1-|sm.west);

  \node[activation] (p1) at (g1-|p0) {$+$};
  \draw[signal] (g1-|sm.east)   -> node[above, midway] {$\cdot\Delta$} (p1);
  \draw[signal] (t0.south west) -> (p1);
  \draw[signal] (p1.east)       -> +(1,0)      node [right] (t1)         {$\vartheta_{1}(x)$};
  \draw[signal, dotted] (t1.south west) -> ($(p1.east)-(0,1)$);

  \node[activation] (pM-1) at (gM-1-|p0) {$+$};
  \draw[signal] (gM-1-|sm.east) -> node[above, midway] {$\cdot\Delta$} (pM-1);
  \draw[signal] (pM-1.east)     -> +(1,0)      node [right] (tM-1)       {$\vartheta_{M}(x)$};
  \draw[signal, dotted] ($(tM-1.north west)+(0,1)-(p1.north)+(p1.south)$) -- (pM-1);

  \node[right] at (O5-|g1) {\raisebox{1.2ex}{$\vdots$}};
  \node at (sm-|p1) {\raisebox{1.2ex}{$\vdots$}};
\end{tikzpicture}
}
  \caption{Determination of the parameters $a_1(\mathbf{x}),b_1(\mathbf{x}),\vartheta_0(\mathbf{x}), \ldots, \vartheta_M(\mathbf{x})$ of the flow $f$ in (\cref{fig:method}) using the unconstrained output of a neural network $\tilde{a}_1(\mathbf{x}),\tilde{b}_1(\mathbf{x}),\tilde\vartheta_0(\mathbf{x}), \ldots, \tilde\vartheta_M(\mathbf{x})$. For illustration purposes a fully connected network is shown, however, any~\gls{NN} architecture can be used that is appropriate for the input modality.}
  \label{fig:NN-para}
\end{figure}

To ensure the invertibility of $f$, we choose the individual transformations to be bijective by requiring strict monotonicity.
For $f_1$ we need to ensure that the scale parameter $a_1$ is positive. This is done by applying a $\softplus$ activation function to the output of the network (\cref{fig:NN-para}).
The required monotonicity of $f_2$ can be easily ensured by enforcing an increasing ordering of the Bernstein coefficients $\vartheta_0,\ldots,\vartheta_M$.

The easiest way to archive this is by recursively applying a strictly positive real function such as $\softplus$ to an unconstrained and unordered vector $\tilde\vartheta_0,\ldots,\tilde\vartheta_M$, such that $\vartheta_0=\tilde\vartheta_0$ and $\vartheta_{k}=\vartheta_{k-1} + \softplus(\tilde\vartheta_k)$ for $k=1,\ldots,M$.
In this simple approach, the convergence in parameter estimation depends on parameter initialization, since $\vartheta_0$ is derived directly from the unconstrained parameters.

We thus require that the transformation $f_2$ covers at least the range $[-3,3]$, i.e., $\pm 3\sigma$ of the standard Gaussian.
Since the boundaries of Bernstein polynomials are given by their first and last coefficient ($f_{2}(0)=\vartheta_{0}$ and $f_{2}(1)=\vartheta_{M}$), we can determine these values from unrestricted parameters $\tilde{\vartheta}_0$ and $\tilde{\vartheta}_{M+1}$ via $\vartheta_0 = -\softplus(\tilde{\vartheta}_0) - 3.0 \le -3$ and $\vartheta_M =  \softplus(\tilde{\vartheta}_{M+1}) + 3.0 \ge 3$.  
To ensure $\sum_{k=1}^M{(\vartheta_k - \vartheta_{k-1})}=\vartheta_M - \vartheta_0 =: \Delta$ the remaining coefficients $\vartheta_k$ for $k=1,\ldots,M$ can be determined as:
\begin{equation}
  \vartheta_k = \vartheta_{k-1} + \Delta\cdot\softmax\left(\left[\tilde\vartheta_1, \tilde\vartheta_3,\ldots, \tilde\vartheta_M\right]\right)_{k-1}
\end{equation}
\noindent
since $\Delta$ and all components of the $\softmax$ are non-negative, $\vartheta_k - \vartheta_{k-1} \ge 0$ so that $f_2$ is monotonous as required.
To sample new data from a learned distribution, the inversion of $f_2$ is required.
Since there is no closed-form solution for the inversion of higher-order Bernstein polynomials, a root-algorithm is used to determine the inverse~\cite{Chandrupatla1997}.

Altogether, the complete transformation $f$ has the following $M+3$ parameters $\theta=\left(a_1(\mathbf{x}), b_1(\mathbf{x}), \vartheta_0(\mathbf{x}), \ldots, \vartheta_M(\mathbf{x})\right)$, obtained from the $M+4$ unconstrained outputs of a~\gls{NN}, as illustrated in~\cref{fig:NN-para}.

Training is done by optimizing the weights of the~\gls{NN}, which controls the parameters of the chained transformations $f_{2} \circ f_1$ w.r.t. the~\gls{NLL} (\cref{eq:nll}) of the training data points, using~\cref{eq:prob}.
All models have been trained with the Adam optimizer~\cite{Kingma2017} and early stopping with a maximum of 300 epochs, to prevent over-fitting.
Additionally, the learning rate was reduced by a factor of ten after every three epochs without significant improvement of the validation loss.

\section{Load Forecasting Simulation Study} \label{sec:study}
This section presents the evaluation of the~\gls{BNF} approach for load forecasting in an empirical study. First, the used dataset is described, then the forecasting approach and benchmark methods are introduced, and finally, the results are discussed.

\subsection{Dataset Description}

The models were trained on a dataset containing information on the electricity demand of smart meter customers in Ireland recorded at a resolution of 30 minutes during the period from 14/07/2009 to 31/12/2010~\cite{CER2012}.
During preprocessing, all non-residential ($2,220$) buildings were dropped, since the stochastic behavior of residential customers ($4,225$) was of explicit interest in this study.
In addition, all incomplete records ($586$) were removed so that the final data set includes $N=3,639$ of the original $6,445$ customers.

To obtain the training data, measurements up to \emph{07/31/2010 23:30} were selected from a random subset of size $N_\text{train}$ of all $N$ households\footnote{the last 10\% of that period were used for validation during development}. This training data is used to determine the weights of the neural network-based approaches.
For testing, there exist two approaches of practical relevance. We can either ask how a model trained on $N_\text{train}$ households performs in the future on the same households (Test 1) or on new, yet unseen, households (Test 3). For completeness, we also include the less relevant case (Test 2), i.e. trained and evaluated in the same period but for different households. An Overview of these data splits is given in~\cref{tab:splits}.

\begin{table}[htb]
  \caption{The dataset~\cite{CER2012} was split by customers and date-ranges into one train and three test sets. Only the Train data was used to optimize the weights of the~\glspl{NN}.}
  \label{tab:splits}
  \centering
  \begin{tabular}{ccc}
                             & \multicolumn{2}{c}{households}            \\
    date range               & used in training               & hold-out \\
    \toprule
    14/07/2009 -- 31/07/2010 & Train                          & Test 2   \\
    01/08/2010 -- 31/12/2010 & Test 1                         & Test 3
  \end{tabular}
\end{table}

The load data has then been normalized into the range $[0,1]$.
Since the exact location of different assets is unknown, no additional data like whether variables have been used.
The scripts used to preprocess the data set and conduct the experiments are available on GitHub\footnote{\url{https://github.com/MArpogaus/stplf-bnf}}.

At runtime, the data is shuffled and batched into mini batches of size 1,024.
Each sample consists of an input tuple $\mathbf{x}=(x_{h}, x_{m})$ containing the \emph{historical} data $x_h$, with the lagged electric load of the past seven days and \emph{meta} data $x_{m}$, with trigonometric encoded time information~\cite[see][]{AndrichvanWyk2018} and a binary holiday indicator as indicated in~\cref{fig:model-scheme}. The prediction target $\mathbf{y}$ is the load for the next day, with resolution of $30$ minutes.
Hence, the models predict $48$ \glspl{CPD} $p(y_1|\mathbf{x}),\ldots,p(y_{48}|\mathbf{x})$ for every future time step.

\subsection{Probabilistic Forecasting Models}\label{sec:model}

\begin{figure}[tbp]
  \resizebox{\columnwidth}{!}{\begin{tikzpicture}[
        scale=2,
        node distance=1.5cm,
        node/.style = {minimum size=2em, font=\sffamily},
        model/.style= {node, draw, fill=lightgray, align = center},
        desc/.style = {pin edge={-,thin,black,dotted}, align=center,font=\em}
    ]
    \node (in) [
        node,
        align=right
    ] {%
    $\begin{aligned}
        x_h&\left|
            \begin{tabular}{p{3.2cm}l}
              load & $\in[0,1]$
            \end{tabular}
            \right.\\
        x_m&\left|
            \begin{tabular}{p{3.2cm}l}
              day of year (sin/cos) & $\in[0,1]$ \\
              weekday (sin/cos) & $\in[0,1]$ \\
              Is holiday & $\in\{0,1\}$
            \end{tabular}
            \right.
    \end{aligned}$
    };
    \node (m) [
        right = of in,
        model,
        pin={[desc]below:{$\theta(x)=NN(\mathbf{x},\omega)$}}
    ] {Network};
    \node (p) [
        right = of m,
        model,
        pin={[desc]below:{$p(\mathbf{y}|\theta(\mathbf{x}))$}}
    ] {Distribution};
    \node (out) [
        node,
        right = of p
    ] {$\begin{array}{c}
      p(y_1|\mathbf{x})\\
      \vdots\\
      p(y_{48}|\mathbf{x})
    \end{array}$};

    \draw[decorate, decoration={brace,amplitude=5pt}] (in.north east) -- (in.south east) node[midway,right=1ex] {$\mathbf{x}$} edge[->] (m);
    \draw (m) -> (p) node[midway,above]{$\theta(\mathbf{x})$};
    \draw[decorate, decoration={brace,amplitude=5pt,mirror}] (out.north west) -- (out.south west) node[midway,left=1ex] {} edge[<-] (p);
\end{tikzpicture}}%
  \caption{Overall model: The variables on the left side are the input to a~\glsxtrfull{NN} (\gls{FC} or~\gls{1DCNN}). The~\gls{NN} controls the parameters $\theta(\mathbf{x})$ for the respective model (\gls{BNF}, \gls{GM}, \gls{GMM}, or  \gls{QR}) yielding the conditional~\gls{CPD} $p(y_{t}|\theta(\mathbf{x}))$ for $48$ time steps.}%
  \label{fig:model-scheme}
\end{figure}
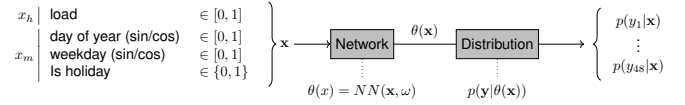

In this study, we used 4 different model types (\gls{BNF}, \gls{GMM}, \gls{GM}, \gls{QR}), which are explained in the following,  to forecast the~\glsxtrfull{CPD} of the load for 48 time steps.
To control the parameters of the forecasting models, we used two~\gls{NN} architectures, a {\em~\glsxtrfull{FC}} and a {\em~\glsxtrfull{1DCNN}} (\cref{fig:model-scheme}).
Architecture choices were made based on prior experience and known best practices~\cite{Geron2019,Chollet2018} and did not undergo further optimizations.

The~\gls{FC} is configured with three hidden layers, with 512, 256 and finally 128 units using the ELU activation function~\cite{Clevert2016}.
The historical data was flattened and concatenated with the metadata.
This model is not specifically designed for processing the sequential data, but serves as a baseline model to assess the influence of a more sophisticated~\gls{NN} model.

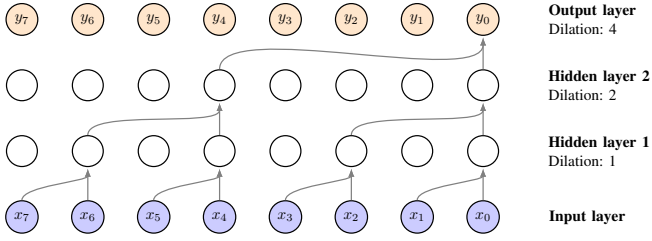
\begin{figure}[htb]
  \centering
  \resizebox{\columnwidth}{!}{\def\layers{4}
\def\inputSize{8}
\def\nodeDist{1.5}
\begin{tikzpicture}
  \foreach \x in {1,...,\inputSize}
    \foreach \y in {1,...,\layers}
    {
      \pgfmathsetmacro{\nodeIdx}{int(\inputSize-\x)}
      \ifx\y\layers%
        \edef\nodeStyle{output node}%
        \edef\nodeText{$y_{\nodeIdx}$}
      \else
        \ifnum\y>1%
          \edef\nodeStyle{hidden node}%
          \edef\nodeText{}
        \else%
          \edef\nodeStyle{input node}%
          \edef\nodeText{$x_{\nodeIdx}$}
        \fi%
      \fi
      \node[%
        \nodeStyle
      ] at (\x*\nodeDist,\y*\nodeDist) (\x\y) {\nodeText};
    }
  \foreach \x in {\inputSize,...,1}{
    \foreach \y in {2,...,\layers}
    {
      \pgfmathsetmacro{\dilation}{int(2^(\y-2))}
      \pgfmathparse{(\x-(\inputSize-2^(\layers-1)) >= \inputSize-2^(\layers-1)&&mod(\x-(\inputSize-2^(\layers-1)),2^(\y-1)) == 0)}
      \ifnum\pgfmathresult>0
        \pgfmathsetmacro{\delaitedX}{int(\x-\dilation)}
        \pgfmathsetmacro{\prevY}{int(\y-1)}
        \draw[%
          signal
        ] (\x\prevY.north) -- (\x\y.south);
        \draw[%
          signal, fill=none
        ] (\delaitedX\prevY.north) to[out=90, in=-90, looseness=.5] (\x\y.south);
      \fi
    }
  }
  \foreach \y in {1,...,\layers} {
    \pgfmathsetmacro{\dilation}{int(2^(\y-2))}
    \ifx\y\layers%
      \def\nodeText{\textbf{Output layer}\\%
      Dilation: \dilation}
    \else%
      \ifnum\y>1%
        \pgfmathsetmacro{\h}{int(\y-1)}
        \def\nodeText{\textbf{Hidden layer \h}\\%
      Dilation: \dilation}
      \else%
        \def\nodeText{\textbf{Input layer}}
      \fi%
    \fi%
    \node[align=left, right = of \inputSize\y] {%
      \nodeText%
    };
  }
\end{tikzpicture}}
  \caption{A stack of dilated causal convolutional layers. The gray arrows indicate the sparse connection to the previous units in the first computational step. Doubling the dilatation rate results in a receptive field of eight, meaning the computation of $y_0$ depends on the input features $x_0\ldots,x_7$. The kernel matrix is moved to the left until all output values are computed. To preserve the input size across the network, zero-padding with the size of the dilation rate is applied to every layer's input. The visualization was inspired by~\cite{Oord2016}.}
  \label{fig:wavenet}
\end{figure}

The~\gls{1DCNN} is inspired by the WaveNet architecture~\cite{Oord2016}. The model was built by stacking $8$ 1D convolutional layers with doubling dilatation rates $1,2,4,\ldots,128$, this principle is illustrated in~\cref{fig:wavenet}.
This results in a model with a receptive field capturing $256$ input values.
Hence, almost the whole input sequence, consisting of one week of historical load data with a total of $348$ input features, can be processed at once.
Each of these dilated convolutions uses the ReLU activation function and has 20 filters.
Finally, a regular 1D convolutional layer without dilatation, again~\gls{ReLU} activation and $10$ filters.
The output of this last convolutional layer is then flattened and concatenated with the metadata before it is fed into a final fully connected layer with~\gls{ELU} activation function, followed by a last layer with a linear activation function to generate the output.

For predicting the~\glspl{CPD} this study proposes the {\em~\glsxtrfull{BNF}} in which $\theta$ are the parameters of the Bernstein polynomial as well as the additional linear transformations making up the flow $f$ (\cref{fig:method}).
We choose $20$ outputs for the~\glspl{NN}, hence the Bernstein polynomials are of order $M=16$.
Higher orders polynomials generally lead to more flexible models, but the improvements are diminishing for high $M$~\cite{Hothorn2018}.
We compare the~\gls{BNF} with three benchmarks to model the~\gls{CPD}, a {\em simple~\glsxtrfull{GM}}, a  {\em~\glsxtrfull{GMM}}, and as a non-parametric approach, a {\em~\glsxtrfull{QR}}. 

The simple~\glsxtrfull{GM} is a probabilistic extension of regular regression, predicting not only the conditioned mean $\mu(x)$, but also the conditional variance $\sigma^2(x)$ of a Gaussian distribution~\cite{Haben2019a, Wijaya2015,Salinas2020}.

To model more complex distribution shapes, e.g., with multiple modes, a~\glsxtrfull{GMM} of three normal distributions was implemented. The output vector $\theta$ contains the mean and variance $\mu_{k}(x),\sigma^2_{k}(x)$ and the mixing coefficients $\alpha_{k}(x)$ for $k=1,2,3$ \citep[see, e.g.,][]{Vossen2018}.

The~\glsxtrfull{QR} is a typical baseline in probabilistic load forecasting~\citep[cf., e.g.,][]{Elvers2019, Wang2019b}. It is configured to predict the $99$ quantiles $p=0.01,\ldots,0.99$ for each time step, which have been constrained to be monotonically increasing by applying a $\softplus$ activation function and then calculating the cumulative sum, to prevent quantile-crossing. Note that the~\gls{QR} is not a continuous~\gls{CPD}, hence the~\gls{NLL} is not tractable, and instead the pinball loss (\cref{eq:mqs}) is minimized.

As a naive baseline, we use the measurements of all households for the respective time-point in the training period to determine the~\glsxtrfull{ECDF}.

\cref{tab:params} summarizes the number of trainable parameters for each model and its output shape.

\begin{table}[ht]
  \caption{Number of trainable parameter and the corresponding output shape for all models}
  \label{tab:params}
  \centering
  \begin{tabular}{llrl}
    \toprule
                           &              & Parameters & Output shape \\
    \gls{NN}               & Distribution &            &              \\
    \midrule
    \multirow{4}{*}{FC}    & BNF          & 463,168    & (48, 20)     \\
                           & GMM          & 395,056    & (48, 9)      \\
                           & GM           & 351,712    & (48, 2)      \\
                           & QR           & 952,336    & (48, 99)     \\
    \midrule
    \multirow{4}{*}{1DCNN} & BNF          & 4,436,794  & (48, 20)     \\
                           & GMM          & 3,895,594  & (48, 9)      \\
                           & GM           & 3,551,194  & (48, 2)      \\
                           & QR           & 8,323,594  & (48, 99)     \\
    \bottomrule
  \end{tabular}
\end{table}

\subsection{Proper Scoring Rules}

Recently, reviews have repeatedly noted that there is a lack of common evaluation methods in the literature on load forecasting, which makes it difficult to compare different methods~\cite{Hong2016a,Hong2020,Haben2021}.
In classical forecasting literature, it is well established to assess probabilistic predictions
in terms of ``\emph{sharpness of the predicted density, subject to calibration}''\citep{Gneiting2007}.
In this context, \emph{calibration} or reliability refers to the statistical consistency between the predicted densities and observations.
\emph{Sharpness} describes the property of predicted density to concentrate tingly around the actual outcome.
In practice, proper scoring rules allow for assessing both sharpness and calibration simultaneously~\citep{Gneiting2014}.
A scoring rule is \emph{proper}, if its value is minimized if the predicted density equals the real distribution and \emph{strictly proper} if this minimum is global, hence only reached if the prediction equals the real distribution~\cite{Gneiting2007}.
We follow these recommendations and evaluate the strictly proper~\gls{NLL} (\cref{eq:nll}) along with two well established proper scoring rules introduced in the following.

The~\glsxtrfull{CRPS} is, besides the NLL, one of the most common proper scores to evaluate the performance of probabilistic forecasts.
It evaluates the quality of the predicted cumulative distribution function and is defined as~\citep{Gneiting2007a,Messner2020}:
\begin{equation}
  \operatorname{CRPS}(F,y)=\frac{1}{N} \sum_{t=1}^{N} \int_{-\infty}^{\infty}\left(F_{t}(x)-\mathbf{1}\left(x \leq y_{t}\right)\right)^{2} d x
\end{equation}
Here $F_{t}$ corresponds to the predicted cumulative density function for time-step $t$ and $\mathbf{1}$ is the unit step function.

To quantify the accuracy of quantile forecasts, often the~\glsxtrfull{MQS} is used with
\begin{equation}\label{eq:mqs}
  \operatorname{MQS}(y_t,q_t,p) = \frac{1}{N} \sum_{t=1}^{N} \left(q_{t}-y_{t}\right)\left(\mathbf{1}\left(y_{t} \leq q_{t}\right)-p\right)
\end{equation}
where $q_{t}$ is the forecast for the $p$-quantile at time-step $t$ where $y_t$ was observed\cite{Messner2020}.
It is equivalent to the Pinball Loss of the~\gls{QR} model~\cite{Wang2019b}.
When predicting continuous distributions or multiple quantiles $\mathbf{q}$, there exists a formulation which can be seen as a discrete variant of~\gls{CRPS} \cite{Laio2007}
\begin{equation}\label{eq:mqs_crps}
  \operatorname{CRPS}(y_t\mathbf{q}_t,\mathbf{p}) \approx 2\Delta p \sum_{\forall q\in \mathbf{q}}\operatorname{MQS}(y,q,p_{q})
\end{equation}
Here the quantile distance $\Delta p$ is assumed constant.
We used the latter, to allow comparability to our~\gls{CRPS} implementation.
Because both~\gls{MQS} and~\gls{CRPS} are scale-dependent, we calculated these metrics using the normalized data for ease of comparison across different datasets and will refer to them as~\glsxtrfull{NMQS} and~\glsxtrfull{NCRPS} and report them as percentages~\cite{Haben2019,VanderMeer2018}.

\subsection{Qualitative Results}

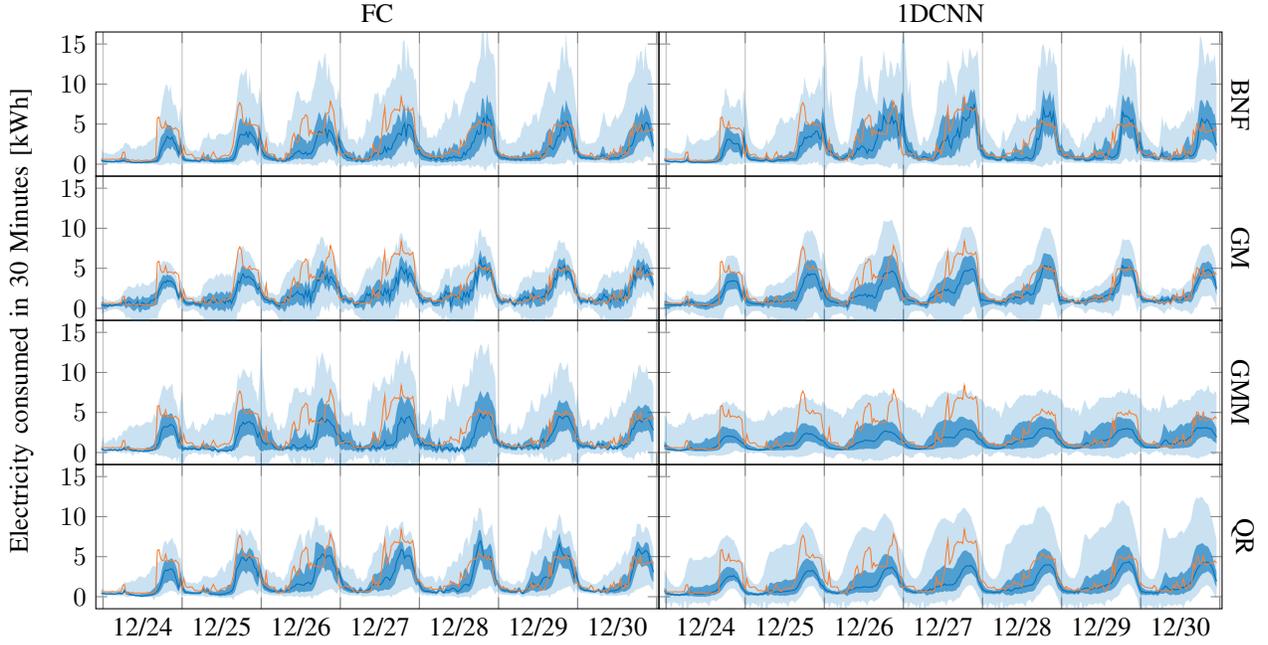
\begin{figure*}[ht!]
  \centering
  \begin{tikzpicture}
  \def\myLabels{}
  \begin{groupplot}[
      group style={
          group size=2 by 4,
          x descriptions at=edge bottom,
          y descriptions at=edge left,
          vertical sep=0pt,
          horizontal sep=0pt,
          every plot/.style = {
              date coordinates in=x,
              xmajorgrids,
              xtick distance = 1,
              x tick label as interval,
              xticklabel={\month/\day},
              legend columns=-1,
              legend style={
                  inner sep=0pt,
                  draw=none,
              },
              every legend to name picture/.style={baseline, inner sep=0pt},
              ymin=0,
              ymax=15,
              enlarge x limits=.01,
              enlarge y limits=.1,
            },
        },
      height=3.5cm,
      width=0.5\textwidth,
    ]
    \def\myPlots{}
    \newcounter{rowCounter}\setcounter{rowCounter}{1}
    \newcounter{colCounter}
    \pgfplotsforeachungrouped \d in {BNF, GM, GMM, QR} {
        \setcounter{colCounter}{1}
        \pgfplotsforeachungrouped \m in {FC, 1DCNN} {
            \edef\nodeName{group c\csuse{thecolCounter}r\therowCounter}
            \ifnum\value{rowCounter}=1
              \xappto{\myLabels}{\noexpand\node[anchor=south] at (\nodeName.north) {\m};}
            \fi
            \ifnum\value{colCounter}=2
              \xappto{\myLabels}{\noexpand\node[anchor=south,rotate=-90] at (\nodeName.east) {\d};}
            \fi
            \eappto\myPlots{%
              \noexpand\nextgroupplot[legend to name=\m_\d_legend_label]
              \noexpand\dateTimeLoadForecastConfidenceAddPlot{\m_\d_christmas}
            }
            \stepcounter{colCounter}
          }
        \stepcounter{rowCounter}
      }
    \myPlots
  \end{groupplot}
  \myLabels
  \node[rotate=90,anchor=south, yshift=2em] at ($(group c1r1.north west)!0.5!(group c1r4.south west)$)
  {Electricity consumed in 30 Minutes [kWh]};
\end{tikzpicture}

  \caption{The plots show the 98\% (\ref{pgfplots:FC_GM_christmas_99}) and 60\% (\ref{pgfplots:FC_GM_christmas_60}) confidence intervals, along with the median (\ref{pgfplots:FC_GM_christmas_Median}) of the predicted~\gls{CPD} and the measured observations (\ref{pgfplots:FC_GM_christmas_Observation}) for one household with unusual high load during the Christmas week. Data from~\cite{CER2012}.}
  \label{fig:confidence_plots_all_models}
\end{figure*}

\cref{fig:confidence_plots_all_models} gives an example of a load forecast for the Irish Christmas holidays for an individual household in Ireland.
The load profile has distinct peaks that correspond to specific activities within the household and exhibits an extraordinarily high consumption compared to the rest of the year.
The plot displays the uncertainty information in the form of confidence intervals for 98\% and 60\%.
This example forecasts in~\cref{fig:confidence_plots_all_models} demonstrates how probabilistic forecasts can handle the specific characteristics of high volatility and complex influences by explicitly modelling the uncertainty, information that is intractable with point forecasting methods.

The two neural networks~\gls{FC} and~\gls{1DCNN} lead to very different results.
In the given examples, the~\gls{FC} variants tend to generate \emph{sharper} predictions, the CI is smaller, and the predicted PDF is concentrating tighter around their mean.
However, the~\gls{FC} models are seemingly less \emph{reliable} and generate more noisy prediction compared to~\gls{1DCNN}, especially for days with unusual high load.
Most models' tails reach far into the negative domain. We observed that our~\gls{BNF} model generates the most ``realistic'' distributions without additional constraints or assumptions about its shape.

\subsection{Dependence on the training data size}

\begin{figure}[htb!]
  \centering
  \includegraphics[width=0.9\columnwidth]{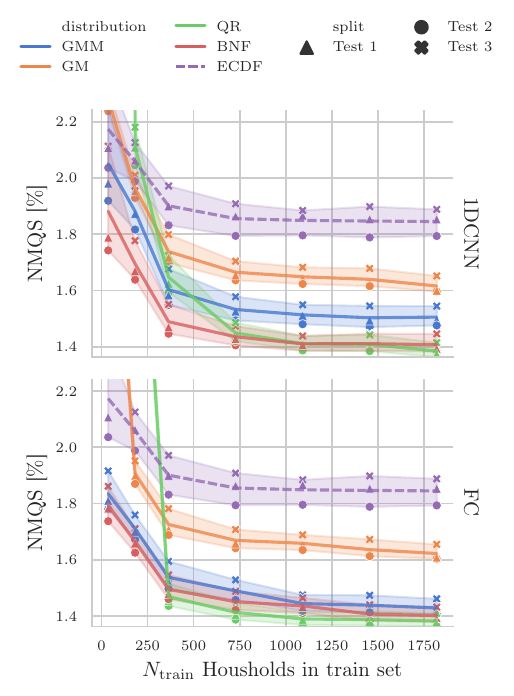}
  \caption{Evaluation results of models trained on a different number of households for $N_\text{train} \in \{36~(1\%)$, $181~(5\%)$, $363~(10\%)$, $727~(20\%)$, $1,091~(30\%)$, $1,455~(40\%)$, $1,819~(50\%)\}$. Shown are the results of~\gls{NMQS} on the three test sets (see~\cref{tab:splits}) along with their mean. The dotted line represents the result of the~\gls{ECDF} baseline model. Lower is better for all values.}
  \label{fig:split_size}
\end{figure}

To assess the performance of the individual models
w.r.t. the number of households $N_\text{train}$ in the training data, we trained all models on different subsets sizes $N_\text{train}$. The results are shown in~\cref{fig:split_size}.

The evaluation results reveal two things.
First, the models' performance appears to asymptotically approach a limit. Beyond a certain point (here about $N_\text{train}=1091~(30\%)$) they do not seem to improve further even when adding more households. Surprisingly, the differences for the individual test sets seams negligible even for small $N_\text{train}$.
It is worth noting that the performance of all models on unknown households in the same period as the training data (Test 2) is always better than on future data (Test 1 and 3)
Predictions of future events for known houses (Test 1) always outperform the prediction for unknown houses (Test 3).
This allows us to conclude that in our case seasonal fluctuations are clearly more relevant than individual behavior patterns, and more households only slightly improve performance on unknown data.
However, it remains unclear whether this is universally true and how future electricity demand will be affected by random events such as electric vehicle charging or rooftop photovoltaics.

Second, the performance of each model in the very data-poor settings $N_\text{train} \le 363~(10\%)$ differs extremely.
For the~\gls{BNF}, \gls{ECDF}, and~\gls{GMM}, we see an almost identical, nearly linear decrease in~\gls{NMQS}, but with unexpectedly good results even for $N_\text{train}$ as low as $36~(1\%)$.
The~\gls{GM} and~\gls{QR} models seem to lose their predictive quality completely and end up with a~\gls{NMQS} outside the plot range. The~\gls{1DCNN} version of~\gls{QR} achieves at best a~\gls{NMQS} of $2,259$.

\subsection{Quantitative Comparison of Models}

\begin{figure*}[htb!]
  \centering
  \includegraphics{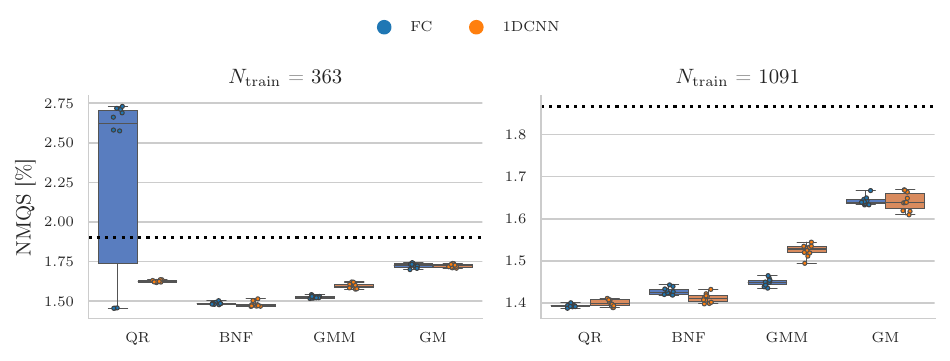}
  \caption{Box plot showing the results of 10 runs with different weight initialization, for all three tests sets (see~\cref{tab:splits}). The dotted line represents the result of the~\gls{ECDF} baseline model. Lower is better for all scores.}
  \label{fig:box_plot}
\end{figure*}

\begin{table}[htb]
  \caption{Test prediction performances achieved with the four~\gls{CPD} models for both~\gls{NN} variants, and the~\gls{ECDF} baseline (\cref{sec:model}).
    The table shows the mean $\mu$ and standard deviation $\sigma$ of the~\gls{NCRPS}, the~\gls{NMQS}, and the~\gls{NLL} of 10 models with different weight initialization, trained on two different sized datasets and evaluated on Test 3.
    Lower is better.
    Note that the~\gls{QR} does not allow us to assess its performance with proper scoring rules, instead, we report the~\gls{NMQS}, for which it was optimized.
    We bold methods if its $\mu + \sigma$ is less or equal to the best model's $\mu - \sigma$}
  \label{tab:res}
  \resizebox{\columnwidth}{!}{%
    \begin{tabular}{llllll}
      \toprule
                               &                           & kind         & NLL                             & NCRPS [\%]                   & NMQS [\%]                    \\
      $N_\text{train}$         & \gls{NN}                  & Distribution &                                 &                              &                              \\
      \midrule
      \multirow[t]{9}{*}{363}  & Baseline                  & ECDF         & -111.702                        & 1.920                        & 1.900                        \\
      \cmidrule{2-6}
                               & \multirow[t]{4}{*}{FC}    & BNF          & \bfseries -135.616 ($\pm$0.388) & \bfseries 1.502 ($\pm$0.007) & \bfseries 1.486 ($\pm$0.007) \\
                               &                           & GMM          & -129.663 ($\pm$0.642)           & 1.542 ($\pm$0.008)           & 1.526 ($\pm$0.008)           \\
                               &                           & GM           & -100.973 ($\pm$0.893)           & 1.743 ($\pm$0.015)           & 1.724 ($\pm$0.014)           \\
                               &                           & QR           & --                              & --                           & 2.303 ($\pm$0.587)           \\
      \cmidrule{2-6}
                               & \multirow[t]{4}{*}{1DCNN} & BNF          & \bfseries -137.040 ($\pm$1.640) & \bfseries 1.495 ($\pm$0.017) & \bfseries 1.479 ($\pm$0.016) \\
                               &                           & GMM          & -132.622 ($\pm$0.560)           & 1.613 ($\pm$0.017)           & 1.596 ($\pm$0.017)           \\
                               &                           & GM           & -100.040 ($\pm$0.408)           & 1.742 ($\pm$0.011)           & 1.724 ($\pm$0.011)           \\
                               &                           & QR           & --                              & --                           & 1.625 ($\pm$0.006)           \\
      \cmidrule{1-6}
      \multirow[t]{9}{*}{1091} & Baseline                  & ECDF         & -114.777                        & 1.886                        & 1.867                        \\
      \cmidrule{2-6}
                               & \multirow[t]{4}{*}{FC}    & BNF          & \bfseries -139.262 ($\pm$0.361) & \bfseries 1.443 ($\pm$0.009) & 1.428 ($\pm$0.009)           \\
                               &                           & GMM          & -135.029 ($\pm$0.754)           & 1.464 ($\pm$0.009)           & 1.449 ($\pm$0.009)           \\
                               &                           & GM           & -104.128 ($\pm$0.402)           & 1.660 ($\pm$0.010)           & 1.642 ($\pm$0.010)           \\
                               &                           & QR           & --                              & --                           & \bfseries 1.393 ($\pm$0.003) \\
      \cmidrule{2-6}
                               & \multirow[t]{4}{*}{1DCNN} & BNF          & \bfseries -142.385 ($\pm$0.904) & \bfseries 1.426 ($\pm$0.011) & \bfseries 1.411 ($\pm$0.011) \\
                               &                           & GMM          & -135.767 ($\pm$0.692)           & 1.541 ($\pm$0.015)           & 1.525 ($\pm$0.014)           \\
                               &                           & GM           & -103.335 ($\pm$1.027)           & 1.659 ($\pm$0.022)           & 1.641 ($\pm$0.021)           \\
                               &                           & QR           & --                              & --                           & \bfseries 1.400 ($\pm$0.008) \\
      \bottomrule
    \end{tabular}
  }
\end{table}

To assess the robustness, all the models have been fitted with 10 different random weight installations for $N_\text{train}\in\{363~(10\%)$, $1,091~(30\%)\}$ customers in the dataset.
The median and standard error for the evaluation on data split \emph{Test 3} (\cref{tab:splits}) are shown in~\cref{tab:res}.
\cref{fig:box_plot} shows the spread for~\gls{MQS} of these individual runs in a box plot.

When looking at the results from~\cref{tab:res} and~\cref{fig:box_plot}, it becomes apparent that the~\gls{GM} model has the worst performance in~\gls{NLL} and~\gls{NCRPS} and more data only slightly improves it.
For the~\gls{NLL} it is even worse than the~\gls{ECDF} baseline.
This illustrates how important it is to pick a density model, which is flexible enough to represent the complex nature of the data.
Interestingly, the scores are not always better the more flexible the~\gls{CPD} is.
We see that the median of~\gls{NMQS} is always lower for the flexible~\gls{BNF} and~\gls{QR} model than for both Gaussian based models~\gls{GM} and~\gls{GMM}, when trained on a $N_\text{train} = 1091$ households.
However, in the low-data scenario, with $N_\text{train} = 363$ \gls{QR} falls behind the~\gls{GMM} for both~\gls{NN} architectures and the~\gls{FC} variant only converges for two of the ten runs.
This confirms our finding that~\gls{QR} requires a solid amount of data to work, and the robustness of~\gls{BNF}.

Furthermore, the~\gls{NN} architecture has an substantial influence on the performance of the~\gls{GMM} and~\gls{QR} models, especially the~\gls{1DCNN} version of the latter performs far better in the low-data setting.
However, the~\gls{BNF} model achieves the overall best scores when trained on $N_\text{train}=363$ households and performs as well as~\gls{QR} for $N_\text{train} = 1,091$.
We speculate that the performance gain of the~\gls{BNF} compared to the~\gls{QR} in the low-data setting is because the~\gls{BNF} is less prone to overfitting since using the Bernstein basis is smoother compared to estimating 99 quantiles, and requires fewer parameters~\cref{tab:params}.
An additional benefit of the~\gls{BNF} over the~\gls{QR} is that it provides a continuous distribution.
This confirms the superiority of the~\gls{BNF} approach, being flexible, stable and requiring less fine-tuning to achieve good results.

\section{Conclusion} \label{sec:conclusion}
Forecasting at the~\gls{LV} level is becoming essential for many stakeholders while more and more applications in low carbon energy systems are explored. Due to the high volatility of load profiles, probabilistic load forecasts are an emerging research topic, as they are capable of expressing uncertainties introduced by the fluctuations caused by the increasing penetration of renewable and distributed energy sources.
The majority of probabilistic load forecasting literature focuses on parametric or~\gls{QR} approaches for estimating marginal distributions.
Parametric methods need to make assumptions about the underlying distribution and so cannot model complex distributions. \gls{QR} based models, on the other hand, can only provide a discrete approximation of the full distribution and thus inherently have the following limitation:
They cannot provide density estimates for arbitrary quantiles, sampling from them is not straightforward, and they do not fit into the maximum likelihood-based deep learning framework.
Instead, the proposed~\gls{BNF} model uses a cascade of parameterized transformation functions, known as~\glsxtrfullpl{NF}, to model the probability density. Model parameters can be obtained by minimizing the~\gls{NLL} through gradient descent and hence fit perfectly into the standard deep learning approach.

We demonstrated that~\glspl{BNF} are a powerful and stable method to express complex non-Gaussian distributions, with
almost no regularization or special tuning.
When there is little data available~\glspl{BNF} performed better than~\gls{QR}, while for larger amounts of training data, both methods are practically en par. This makes~\glspl{BNF} a preferential choice over the~\glspl{QR} or Gaussian approaches for probabilistic load forecasts.

\begin{figure}
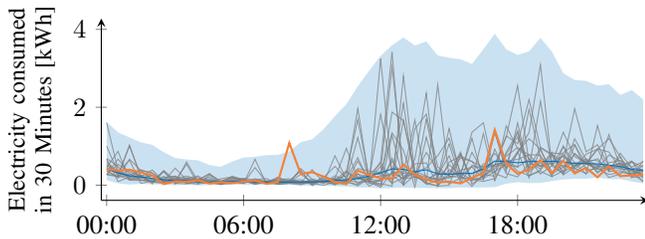

  \centering
  \samplePlot[%
    height=4cm,
    width=\columnwidth,
    ymin=0,
    enlarge x limits=.01,
    enlarge y limits=.1,
    legend to name=sample_legend_label,
    ylabel={Electricity consumed\\in 30 Minutes [kWh]}
  ]{samples}{15}
  \caption{Our~\gls{BNF} approach allows sampling from the learned distributions.
    The plot shows the 99\% (\ref{pgfplots:samples_99}) confidence intervals, median (\ref{pgfplots:samples_Median}), the observed values (\ref{pgfplots:samples_Observation}), and 15 samples (\ref{pgfplots:samples_Samples}) drawn from the predicted~\gls{CPD}.}
  \label{fig:sampling}
\end{figure}

A possible enhancement might be to take the multivariate nature of the forecast more directly into account. Instead of predicting multiple marginal~\glspl{CPD} for a fixed forecast horizon, a future implementation could benefit from 
autoregressive architectures for non-fixed forecast horizons or extending the BNF to \mbox{multivariate versions} and use it, i.e., for sampling new realistic load profiles (see exemplarily in~\cref{fig:sampling}).
This makes~\glspl{BNF} also applicable for other use cases like generation of synthetic scenarios for grid planning or stochastic optimization approaches.
In addition, ensembles have demonstrated superior predictive performance in probabilistic load forecasting\cite{Wang2019c}, thus recent transformation ensembles\cite{Kook2022a} are a promising extension of the current framework to improve both performance and interpretability.

\section*{Acknowledgments}
Parts of this work have been funded by the Federal Ministry for the Environment, Nature Conservation and Nuclear Safety due to a decision of the German Federal Parliament (AI4Grids: 67KI2012A), by the Federal Ministry for Economic Affairs and Energy (BMWi) within the program SINTEG as part of the showcase region WindNODE (03SIN539) and by the Federal Ministry
of Education and Research of Germany (BMBF) in the project
DeepDoubt (grant no. 01IS19083A).
Public data from the CER Smart Metering Project accessed via the Irish Social Science Data Archive (ISSDA) were used in the development of this project~\cite{CER2012}.

The authors would like to thank the reviewers for their valuable and detailed comments.
\printbibliography

\vfill\break

\begin{IEEEbiography}[{\includegraphics[width=1in,height=1.25in,clip,keepaspectratio,trim=6 0 7 0]{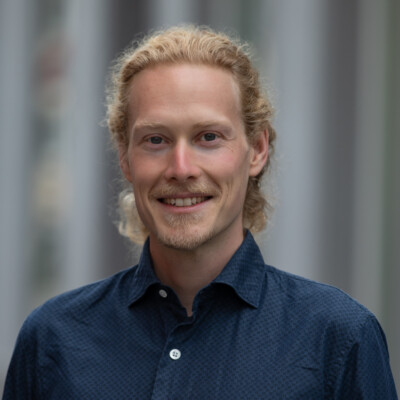}}]{Marcel Arpogaus}
  holds a B.\,Eng. in Electrical Engineering and received his M.\,Sc. degree in Computer Science from the Konstanz University of Applied Science (UAS) in 2020.

  He is currently working as a Research Associate in the research project ``AI4Grids'' at UAS, where he is developing machine and deep learning algorithms for energy forecasting, distribution grid operation and planning.
\end{IEEEbiography}
\vskip -2\baselineskip
\begin{IEEEbiography}[{\includegraphics[width=1in,height=1.25in,clip,keepaspectratio,trim=20 0 22 0]{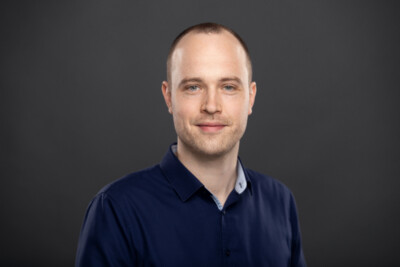}}]{Marcus Voss} received the M.\,Sc. degree in Business Information Systems from the Humboldt University of Berlin in 2014.

  He is an AI expert at Birds on Mars GmbH and Ph.\,D. student at TU Berlin. Prior he was a Research Associate at the Distributed Artificial Intelligence Laboratory
  of the TU Berlin, leading the research group "Smart Energy Systems".
  In his Ph.\,D. research, he focuses on analyzing low-voltage smart meter data using machine learning.
\end{IEEEbiography}
\vskip -2\baselineskip
\begin{IEEEbiography}[{\includegraphics[width=1in,height=1.25in,clip,keepaspectratio]{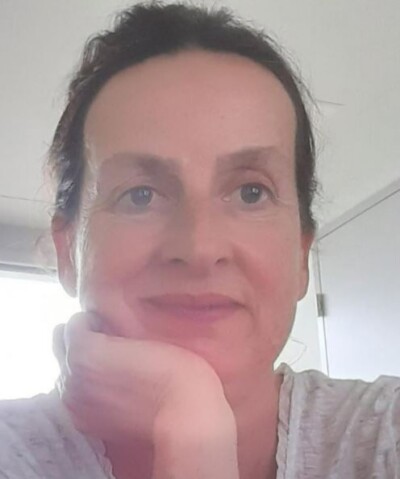}}]{Beate Sick}
  Beate Sick is a professor for applied statistics at Zurich University of Applied Science and co-affiliated at University Zurich. After her PhD in experimental physics at ETHZ, she turned to biostatistics and was responsible for the bioinformatics at the DNA Array facility of UNIL and EPFL before joining ZHAW. Currently, her main research focus is on leveraging deep learning approaches for medical research.
\end{IEEEbiography}
\vskip -2\baselineskip
\begin{IEEEbiography}[{\includegraphics[width=1in,height=1.25in,clip,keepaspectratio,trim=6 0 7 0]{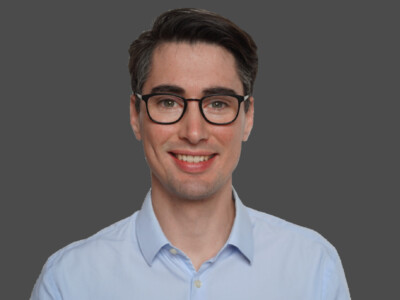}}]{Mark Nigge-Uricher}
  received the M.Sc. degree in Global Innovation Management from the Strathclyde and Aalborg University in 2012.

  He is a Senior Managing Consultant with Bosch.IO GmbH. There he was responsible for the low voltage energy management domain, mainly related to Virtual Power Plants. Additionally, he  consults in IoT and Business Model Innovation projects.
\end{IEEEbiography}
\vskip -2\baselineskip
\begin{IEEEbiography}[{\includegraphics[width=1in,height=1.25in,clip,keepaspectratio]{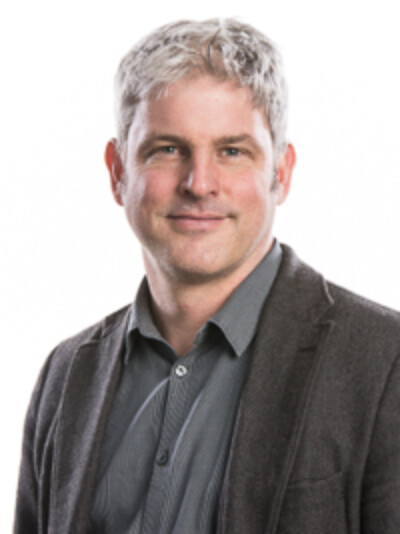}}]{Oliver Dürr}
  is a professor for data science at the Konstanz University of Applied Sciences. After his Ph.D. in theoretical physics, he worked 10 years in a bioinformatics company developing and applying machine learning and statistical methods to all kinds of -omics data. He then was a lecture for statistical data analysis at the Zurich University of Applied Sciences. He is now working on deep learning mainly from a probabilistic perspective.
\end{IEEEbiography}

\vfill

\end{document}